\documentclass[letterpaper]{article} 
\usepackage{aaai2026}  
\usepackage{times}  
\usepackage{helvet}  
\usepackage{courier}  
\usepackage[hyphens]{url}  
\usepackage{graphicx} 
\usepackage{natbib}  
\usepackage{caption} 
\usepackage{algorithm}

\usepackage{newfloat}
\usepackage{listings}

\usepackage{multicol}

\usepackage{booktabs}
\usepackage{enumitem}
\usepackage{amsmath}
\usepackage[noend]{algpseudocode}  
\usepackage{multirow}
\usepackage{adjustbox}
\usepackage{colortbl}
\usepackage{subfigure}
\usepackage{float}
\usepackage[table]{xcolor}
\usepackage{amssymb}

\DeclareCaptionStyle{ruled}{labelfont=normalfont,labelsep=colon,strut=off} 
\floatstyle{ruled}
\newfloat{listing}{tb}{lst}{}
\floatname{listing}{Listing}
\nocopyright

\title{NAS-LoRA: Empowering Parameter-Efficient Fine-Tuning for Visual Foundation Models with Searchable Adaptation}
\author{
    Renqi Chen\textsuperscript{\rm 1},
    Haoyang Su\textsuperscript{\rm 2},
    Shixiang Tang\textsuperscript{\rm 3}\thanks{Corresponding author.}
}
\affiliations{
    \textsuperscript{\rm 1}College of Intelligent Robotics and Advanced Manufacturing, Fudan University, Shanghai, China\\
    \textsuperscript{\rm 2}{Australian Institute for Machine Learning, The University of Adelaide, Adelaide, Australia}\\
    \textsuperscript{\rm 3}{Shanghai Artificial Intelligence Laboratory}

    rqchen23@m.fudan.edu.cn, haoyang.su@adelaide.edu.cn, tangshixiang@pjlab.org.cn
}

\usepackage{bibentry}

\begin{document}

\maketitle

\begin{abstract}
The Segment Anything Model (SAM) has emerged as a powerful visual foundation model for image segmentation. However, adapting SAM to specific downstream tasks, such as medical and agricultural imaging, remains a significant challenge. To address this, Low-Rank Adaptation (LoRA) and its variants have been widely employed to enhancing SAM's adaptation performance on diverse domains. Despite advancements, a critical question arises: can we integrate inductive bias into the model? This is particularly relevant since the Transformer encoder in SAM inherently lacks spatial priors within image patches, potentially hindering the acquisition of high-level semantic information.
In this paper, we propose NAS-LoRA, a new Parameter-Efficient Fine-Tuning (PEFT) method designed to bridge the semantic gap between pre-trained SAM and specialized domains.
Specifically, NAS-LoRA incorporates a lightweight Neural Architecture Search (NAS) block between the encoder and decoder components of LoRA to dynamically optimize the prior knowledge integrated into weight updates. Furthermore, we propose a stage-wise optimization strategy to help the ViT encoder balance weight updates and architectural adjustments, facilitating the gradual learning of high-level semantic information. Various Experiments demonstrate our NAS-LoRA improves existing PEFT methods, while reducing training cost by \textbf{24.14\%} without increasing inference cost, highlighting the potential of NAS in enhancing PEFT for visual foundation models.
\end{abstract}

\vspace{-2mm}
\section{Introduction}

Foundation models pre-trained on large-scale general datasets have demonstrated remarkable generalization capabilities across diverse tasks~\cite{touvron2023llama,liu2023visual,chen2024internvl}. Among visual foundation models, the Segment Anything Model (SAM)~\cite{kirillov2023segment} stands out for its prompt-based design and training on over a billion masks, {achieving strong zero-shot performance on natural images and common objects}. However, SAM struggles with domain-specific tasks such as medical imaging and remote sensing~\cite{chen2023sam,wu2023medical,osco2023segment}, where domain knowledge and visual inductive biases are critical~\cite{wu2023medical,chen2024ma}.
While fully fine-tuning foundation models for such tasks is often computationally intensive and prone to overfitting, Low-Rank Adaptation (LoRA)~\cite{hu2022lora}, which is a Parameter-Efficient Fine-Tuning (PEFT)~\cite{houlsby2019parameter,tu2023visual,xin2024vmt,wang2024lion} method, is proposed to address these challenges by updating only a small subset of parameters. Despite advancements, its capacity to encode domain knowledge and inductive biases remains underexplored, {partly due to the ViT encoder's lack of inherent task-specific priors. Unlike convolutional architectures, ViTs process input globally without built-in locality or structure-aware mechanisms, making it harder to capture fine-grained patterns and contextual cues essential for specialized domains such as medical or remote sensing imagery~\cite{zhong2024convolution}.}


\begin{figure}[t]
  \centering
  \subfigure[LoRA vs. NAS-LoRA]{\includegraphics[width=0.44\linewidth]{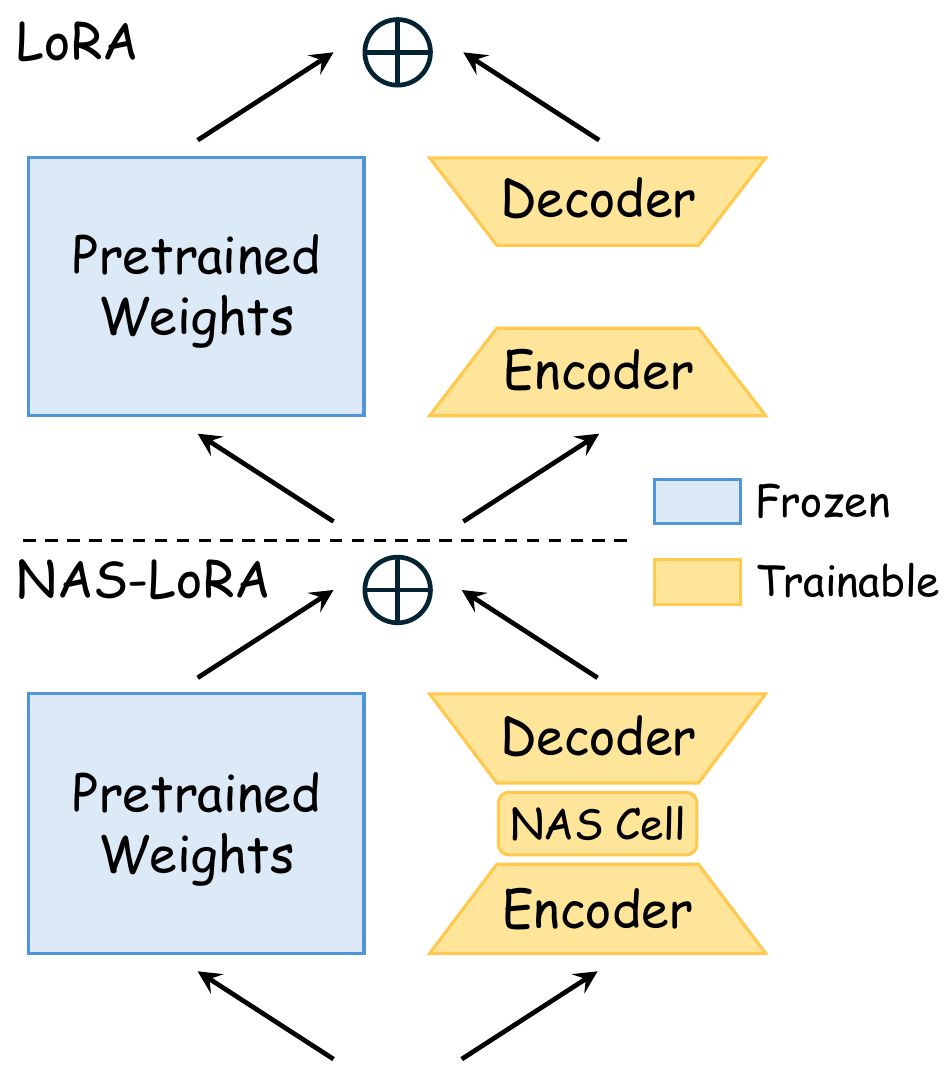}}
  \hspace{-2mm}
  \subfigure[Comparison on various datasets]{\includegraphics[width=0.56\linewidth]{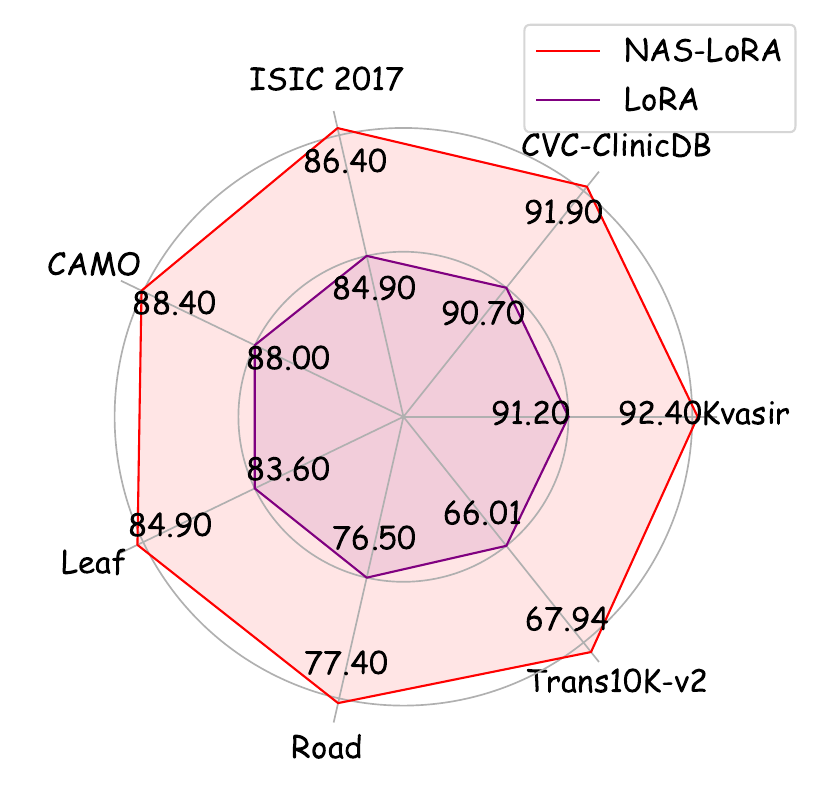}}
  \vspace{-2mm}
 \subfigure[Training cost and inference time]{\includegraphics[width=0.97\linewidth]{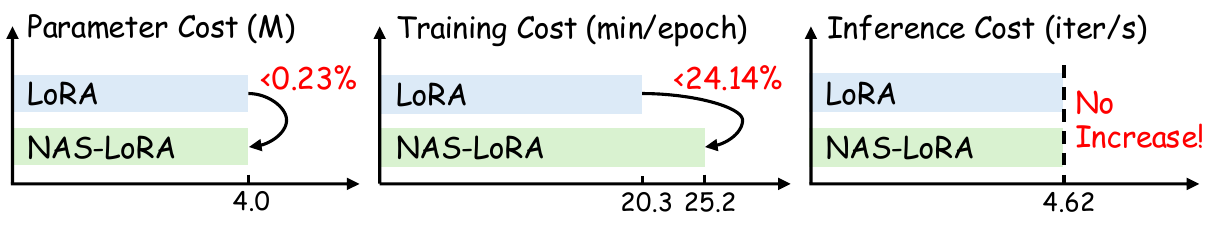}}
 \vspace{-3mm}
  \caption{NAS-LoRA adds a lightweight, trainable NAS cell between LoRA's encoder and decoder to inject optimized prior knowledge. This enhancement shows superiority across each segmentation tasks, with only a minimal increase in training cost and no additional inference overhead.}
  \label{fig:overview}
\end{figure} 


{To address the lack of local priors in SAM, convolutional LoRA~\cite{zhong2024convolution} introduces convolutions into LoRA modules across all encoder layers. However, operating on $Q$, $K$, and $V$ embeddings rather than raw pixels may distort spatial structure. Moreover, critical components like pooling and skip connections~\cite{ronneberger2015u,he2016deep,chen2021learning,nirthika2022pooling} are absent in such adaptations. This raises a key question: Should convolution be applied uniformly, or can we design a more adaptive way to inject inductive biases?}

{We argue that different downstream tasks often require tailored inductive biases. For instance, cancer image segmentation relies on capturing fine-grained textures and subtle intensity variations to delineate ambiguous boundaries~\cite{chen2023transnorm,wang2022dofa}, while natural image segmentation benefits more from strong edge detection and global spatial context~\cite{kirillov2023segment,zhang2023segformerv2}. Similarly, remote sensing tasks emphasize geometric regularity and scale invariance to detect structures like roads or buildings under varying resolutions. These differences highlight the need for flexible adaptation strategies that can selectively inject task-relevant priors, rather than applying a one-size-fits-all design across domains.}

Inspired by Neural Architecture Search (NAS)~\cite{liu2018darts,xu2019pc,salehin2024automl}, which can dynamically optimizes network structures, we propose NAS-LoRA, an adaptive PEFT framework that generalizes SAM across diverse downstream tasks.
As shown in Fig.~\ref{fig:overview}, NAS-LoRA integrates a lightweight NAS cell between the encoder and decoder of LoRA, 
{enabling dynamic selection of the most suitable feature transformation pathways for each task.}
{To avoid sub-optimal results caused by complex search spaces and parameter-architecture interactions, we introduce a stage-wise strategy to decouple the optimization process, ensuring gradual semantic learning in the ViT encoder.}
Additionally, unlike traditional NAS methods requiring a decoding step for final structure selection~\cite{liang2021efficient,wang2024advances,salmani2025systematic}, NAS-LoRA seamlessly merges into pre-trained weights without re-training or merging loss, maintaining its efficiency. {We evaluate NAS-LoRA across nine segmentation benchmarks from different domains, showing its superiority over existing PEFT methods. Additionally, NAS-LoRA outperforms LoRA by reducing training cost by 24.14\% without increasing inference cost.}
Our key contributions are summarized as follows:

\begin{itemize}
    \item We introduce NAS-LoRA, a novel PEFT framework that dynamically selects optimal pathways for inductive bias injection during weight updates, effectively adapting SAM to specialized domains.
    \item We propose a a stage-wise optimization strategy that balances weight adaptation and architectural adjustments, improving high-level semantic learning in SAM’s ViT encoder.
    \item NAS-LoRA eliminates the need for additional decoding and re-training steps, unlike traditional NAS approaches, ensuring a parameter-efficient adaptation process.
\end{itemize}
\vspace{-2mm}
\section{Related work}
\noindent \textbf{Parameter Efficient Fine-Tuning.}
To reduce computational costs and mitigate overfitting, Parameter Efficient Fine-Tuning (PEFT) methods adapt pre-trained models to downstream tasks by fine-tuning only a small subset of parameters while keeping the original model weights frozen~\cite{han2024parameter,xin2024parameter}. PEFT techniques can be broadly categorized into adapter-based methods, prompt-based tuning, and Low-Rank Adaptation (LoRA).
Adapter-based methods~\cite{houlsby2019parameter,yin20231,yang2023aim,yin20255} introduce additional trainable modules between the layers of a pre-trained model, while prompt-based tuning~\cite{lester2021power,razdaibiedina2023residual,wang2024lion} appends learnable tokens to the input and fine-tunes only these tokens. Despite their effectiveness, both approaches face two main limitations: (1) sensitivity to the initialization of learnable parameters and (2) increased inference latency due to added modules or input modifications.

LoRA~\cite{hu2022lora} overcomes these issues by introducing low-rank matrices to approximate weight updates, which are merged into the pre-trained model before inference, avoiding additional computational overhead. LoRA variants~\cite{kopiczko2023vera,liu2024dora,zhong2024convolution} have been proposed to analyze its mechanism and enhance its effectiveness. For instance, DoRA~\cite{liu2024dora} leverages weight decomposition analysis to reveal that the gap between LoRA and full fine-tuning is not solely due to the number of trainable parameters but the patterns of weight updates. Instead of increasing parameter efficiency through weight decomposition, we focus on optimizing the arrangement of trainable parameters within LoRA. 
This design allows LoRA to incorporate not only domain-specific knowledge but also structured inductive biases, improving its adaptation to downstream tasks.

\begin{figure*}[ht]
  \centering
  \includegraphics[width=0.92\linewidth]{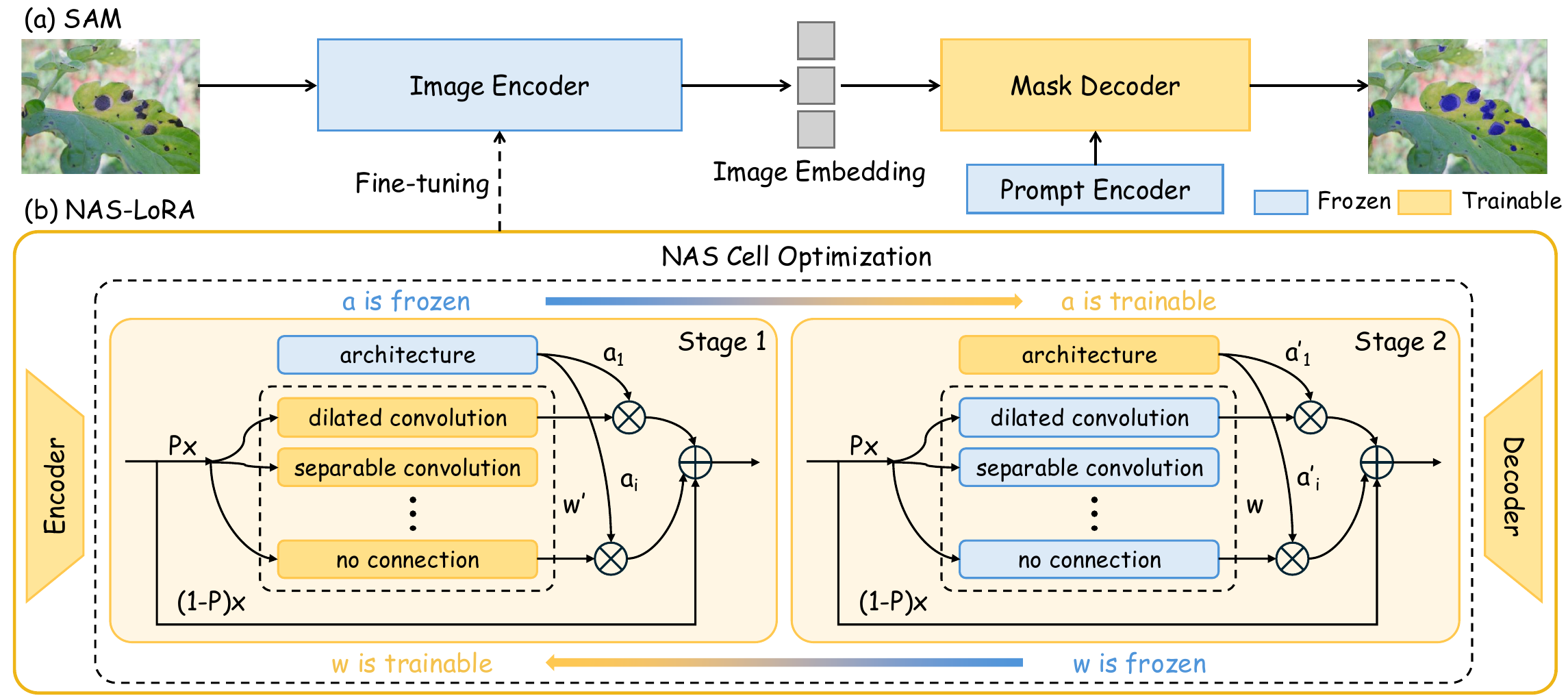}
  \vspace{-1mm}
  \caption{The proposed NAS-LoRA framework for fine-tuning SAM. The upper part illustrates the design of each SAM component: NAS-LoRA is applied to the self-attention layers of the image encoder, the prompt encoder is frozen to enable automated processing, and the mask decoder is fully fine-tuned without LoRA, as it is a lightweight module. The lower part depicts the end-to-end stage-wise optimization of NAS-LoRA, where Stage 1 and Stage 2 iteratively update model weights and architecture parameters using two independent optimizers. After optimization, the learned architecture parameters and weights are directly merged into the pre-trained model for downstream tasks, following the standard LoRA merging process.}
  \label{fig:nas-lora}
\end{figure*}

\noindent \textbf{Fine-tuning SAM.}
Segment Anything (SAM)~\cite{kirillov2023segment}, a foundational model trained on the SA-1B dataset with over one billion masks using a model-in-the-loop approach, has demonstrated strong generalization and zero-shot segmentation capabilities. Given its zero-shot performance, existing studies have integrated LoRA or inserted adapters to enhance SAM’s segmentation capabilities for specific visual tasks~\cite{chen2023sam,wu2023medical,zhang2023customized,zhong2024convolution}. Among these, Conv-LoRA~\cite{zhong2024convolution} represents a notable attempt to incorporate inductive biases by injecting convolutional operations between the encoder and decoder components of LoRA. While this modification improved performance compared to standard LoRA fine-tuning, it applies convolutions uniformly across all SAM encoder layers, disregarding the structural differences in patch embeddings and their varying sensitivity to local priors. To address this limitation, we propose NAS-LoRA, which dynamically selects the optimal feature mapping pathway for prior incorporation. By adaptively injecting inductive biases, NAS-LoRA achieves a more efficient and effective fine-tuning strategy.

\noindent \textbf{Neural Architecture Search.}~
Neural Architecture Search (NAS) aims to discover optimal neural architectures while minimizing computational costs and reliance on expert knowledge~\cite{ren2021comprehensive}. To enhance efficiency, gradient-based NAS methods~\cite{liu2018darts,xu2019pc,chu2021fairnas,ye2022b,he2024darts} have been introduced, leveraging continuous relaxation to reallocate path weights across candidate operations within layers or cells. Additionally, efforts to streamline NAS~\cite{xu2019pc,xue2022partial,gao2022rf} focus on reducing the computational burden associated with the search process.

\noindent \textbf{NAS vs. NAS-LoRA.}~However, integrating the advancements of conventional NAS methods into foundation model during fine-tuning exists two critical challenges. 
First, they typically require a decoding step—such as the Viterbi algorithm—to finalize the selected architecture, which introduces biases and necessitate additional re-training to align the decoded model with the target task. Second, NAS is computationally intensive during training, making it impractical for large-scale models.
To overcome these, we propose NAS-LoRA, which eliminates both decoding and re-training by directly optimizing weight updates, allowing seamless integration with pre-trained models. Furthermore, by utilizing the low rank of LoRA and channel selection, NAS-LoRA introduces a lightweight search framework, efficiently optimizing the allocation of trainable parameters within LoRA. 

By appropriately applying NAS principles to large-scale models, NAS-LoRA achieves improvements in adaptation without incurring excessive computational costs, making it the first scalable and practical NAS-based solution for fine-tuning foundation models.


\vspace{-0.5mm}
\section{Methodology}
\subsection{Preliminary: LoRA}
As shown in the top section of Fig.~\ref{fig:overview}(a), LoRA~\cite{hu2022lora} fine-tunes a model by freezing the pre-trained weights and introducing a low-rank adaptation mechanism. Specifically, LoRA injects a pair of trainable low-rank matrices (an encoder and a decoder) into each transformer layer, enabling efficient weight updates with minimal additional parameters. Formally, given a pre-trained weight matrix $W_{0}\in\mathbb{R}^{b\times a}$, LoRA introduces an encoder matrix $W_{e}\in\mathbb{R}^{r\times a}$ and a decoder matrix $W_{d}\in\mathbb{R}^{b\times r}$, where $r\ll min(a,b)$. The forward pass with LoRA is defined as:
\begin{equation}
    h = W_{0}x+\Delta Wx = W_{0}x+W_{d}W_{e}x
\end{equation}

In contrast, NAS-LoRA, shown in the bottom section of Fig.~\ref{fig:overview}(a), extends this framework by incorporating a trainable NAS cell between the encoder and decoder. Unlike Conv-LoRA\cite{zhong2024convolution}, which injects inductive biases by applying convolutions indiscriminately across all transformer layers, NAS-LoRA introduces a dynamic selection mechanism. This mechanism adaptively determines the optimal feature mapping pathway between LoRA’s encoder and decoder, effectively bridging the semantic gap between pre-trained models and downstream tasks. By optimizing inductive bias injection, NAS-LoRA enables more adaptive and efficient fine-tuning, ensuring that local priors are incorporated in a task-specific manner.
\vspace{-2mm}
\subsection{NAS-LoRA}
NAS-LoRA introduces a NAS cell that dynamically selects the optimal local prior. 
As illustrated in Fig.\ref{fig:nas-lora}(b), the NAS cell evolves across different training stages (discussed in Sec.\ref{sec:optim}) and serves as a learnable module that adaptively determines the optimal feature mapping pathway. This pathway is constructed from a set of candidate operations $\mathcal{O}$, with the most suitable operation being selected based on architecture parameters $\alpha$. In this work, $\mathcal{O}$ comprises widely used operations in modern CNNs, including:
\begin{minipage}{0.59\linewidth}
  \small
  \medbreak
  \begin{itemize}[leftmargin=*, nolistsep]
  \item $3\times3$ separable convolution
  \item $5\times5$ separable convolution
  \item $3\times3$ convolution \& dilation rate $2$
  \item $5\times5$ convolution \& dilation rate $2$
  \end{itemize}
  \medbreak
  \end{minipage}
  \begin{minipage}{0.39\linewidth}
  \small
  \medbreak
  \begin{itemize}[leftmargin=*, nolistsep]
  \item $3\times3$ average pooling
  \item $3\times3$ max pooling
  \item skip connection
  \item no connection (zero)
  \end{itemize}
  \medbreak
  \end{minipage}
  
Each operation serves a distinct function: convolution captures local features and enables feature transformation, pooling reduces spatial dimensions and assists in the acquisition of invariant features, skip connection enhances gradient flow and facilitates information fusion, while no connection enables information flow isolation.

The forward pass of NAS-LoRA is formulated as:
\vspace{-2mm}
\begin{equation}\label{eq:NAS-LoRA}
    h = W_{0}x+W_{d}(\sum_{i=1}^{|\mathcal{O}|} \frac{exp\{\alpha_{i}\}}{\sum_{j=1}^{|\mathcal{O}|} exp\{\alpha_{j}\}}O_{i}(W_{e}x)),
\end{equation}
where $W_{0}\in\mathbb{R}^{C_{out}\times C_{in}}$, $W_{e}\in\mathbb{R}^{r\times C_{in}}$, $W_{d}\in\mathbb{R}^{C_{out}\times r}$, and $x\in\mathbb{R}^{B\times C_{in}\times H\times W}$. 
Here, $C_{in}$ and $C_{out}$ denote the number of input and output channels, $r$ is the rank of LoRA, $B$ is the batch size, $H$ and $W$ are the input dimensions. $O_{i}$ represents the $i$-th operation in $\mathcal{O}$, and the architecture parameter $\alpha$ is optimized by the continuous relaxation algorithm. This formulation ensures that the most effective prior is incorporated into the parameter updates, enabling a parameter-efficient NAS search.

\noindent \textbf{NAS with PEFT.} 
Although the NAS cell can autonomously adjust its structure and parameters to enhance representation learning, its built-in search process adds computational overhead, undermining the efficiency goals of PEFT.
Therefore, maintaining the lightweight and efficient nature of NAS-LoRA emerges as a critical challenge.
To address this, 
we take inspiration from Partial Connections~\cite{xu2019pc}, which show that training with a subset of feature channels maintains performance without degradation.
By integrating this \textbf{P}artial \textbf{C}onnection mechanism into NAS-LoRA, we introduce NAS-\textbf{PC}-LoRA, where Eq.~\ref{eq:NAS-LoRA} is reformulated as:
\vspace{-2mm}
\begin{equation}
\resizebox{\linewidth}{!}{$
h = W_{0}x + W_{d}\left[(1-\mathbf{P})\odot(W_{e}x) +
\sum_{i=1}^{|\mathcal{O}|} \frac{\exp\{\alpha_{i}\}}{\sum_{j=1}^{|\mathcal{O}|} \exp\{\alpha_{j}\}} O_{i}(\mathbf{P}\odot W_{e}x)\right]
$}
\end{equation}
where $\mathbf{P}$ is a binary mask indicating the selected channels, and $\odot$ denotes the Hadamard product. Notably, when $\mathbf{P}$ is a zero matrix, NAS-PC-LoRA degrades to standard LoRA.
\vspace{-2mm}
\subsection{NAS-LoRA in Training}
\label{sec:optim}
As illustrated in Fig.~\ref{fig:nas-lora}(a), we adopt SAM as the backbone for NAS-LoRA, which consists of three key components: an image encoder, a mask decoder, and a prompt encoder. For the image encoder, we integrate NAS-LoRA along with a stage-wise optimization strategy to effectively adapt the model to downstream tasks (detailed in Alg.~\ref{alg:NAS-LoRA}). The mask decoder is fully fine-tuned without LoRA. To enable end-to-end model optimization and application, we freeze the prompt encoder while maintaining constant prompt tokens. Notably, the original mask decoder is modified from binary mask prediction to multi-class mask prediction by incorporating a classification branch. This branch is responsible for predicting classification scores~\footnote{Following the design in Conv-LoRA~\cite{zhong2024convolution} and Mask2Former~\cite{cheng2022masked}, the classification branch produces $K+1$ class predictions (with an additional category for 'no object'), assuming the original task has $K$ categories. For semantic inference, pixel-wise predictions are obtained by performing a matrix multiplication between the masks and classification predictions, followed by an additive operation.}.

\noindent \textbf{Loss Functions.}~
Our final loss function combines the segmentation loss $\mathcal{L}_{\text{seg}}$ and classification loss $\mathcal{L}_{\text{cls}}$ as follows: 
\begin{equation} 
\mathcal{L} = \lambda_{\text{seg}} \mathcal{L}_{\text{seg}} + \lambda_{\text{cls}} \mathcal{L}_{\text{cls}}, 
\end{equation} 
where $\lambda_{\text{seg}}$ and $\lambda_{\text{cls}}$ are the weighting coefficients that balance the two objectives.
The segmentation loss $\mathcal{L}_{\text{seg}}$~\cite{ma2024segment} includes a binary cross-entropy (BCE) loss $\mathcal{L}_{\text{BCE}}$ and a Dice loss $\mathcal{L}_{\text{Dice}}$.
The classification loss $\mathcal{L}_{\text{cls}}$ is defined as the cross-entropy over all $K+1$ classes. Further details about the loss functions are shown in Appx.~\ref{sec:app_loss}.




\noindent \textbf{Optimization Strategy.}~As shown in Fig.~\ref{fig:nas-lora}(b), stage-wise optimization strategy follows a two-stage iterative process, where architecture parameters and model weights are updated separately using two independent optimizers. The procedure is shown in Alg.~\ref{alg:NAS-LoRA}.
\vspace{-0.6mm}
\begin{algorithm}
  \caption{NAS-LoRA Optimization}
  \label{alg:NAS-LoRA}
  \begin{algorithmic}[1]
    \State \textbf{Input:} weight parameters $w$, architecture parameters $\alpha$, optimization steps $T$, and $T_B$
    \State \textbf{Output:} Optimized $w', \alpha'$
    \For{$t = 1$ to $T$}
      \State Fix $\alpha$ and update $w$ by $\nabla_{w}\mathcal{L}(w,\alpha)$ \hfill$\longrightarrow$ \textbf{Stage 1}
      \If{$t > T_B$}
        \State Fix $w$ and update $\alpha$ by $\nabla_{\alpha}\mathcal{L}(w,\alpha)$ \hfill$\longrightarrow$ \textbf{Stage 2}
      \EndIf
    \EndFor
  \end{algorithmic}
\end{algorithm}
\vspace{-2.5mm}
\subsection{NAS-LoRA in Inference}
In LoRA, the fine-tuned weights are merged with the frozen pre-trained weights as follows:
\begin{equation}
h=W_{0}x+W'_{d}W'_{e}x=\underbrace{(W_{0}+W'_{d}W'_{e})}_{W_{merged}}x.
\end{equation}

Similarly, in NAS-LoRA, the merging process remains straightforward. The weighted combination of candidate operation weights and architecture parameters can be directly integrated into the pre-trained weights, which can be formulated based on Eq.~\ref{eq:NAS-LoRA} as:
\begin{equation}
  \begin{aligned}
    h&=\underbrace{(W_{0}+W'_{d}(\sum_{i=1}^{|\mathcal{O}|} \frac{exp\{\alpha'_{i}\}}{\sum_{j=1}^{|\mathcal{O}|} exp\{\alpha'_{j}\}}O_{i}W'_{e}))}_{W_{merged}}x.
  \end{aligned}
\end{equation}

For NAS-PC-LoRA, the merged weight is formulated as:
\begin{equation}
\resizebox{\linewidth}{!}{$
  W_{merged}=W_{0}+W'_{d}[(1-\mathbf{P})\odot W'_{e}+\sum_{i=1}^{|\mathcal{O}|} \frac{exp\{\alpha'_{i}\}}{\sum_{j=1}^{|\mathcal{O}|} exp\{\alpha'_{j}\}}O_{i}(\mathbf{P}\odot W'_{e})].
  $}
\end{equation}

A key advantage of NAS-LoRA is that, unlike traditional NAS, it eliminates the need for an additional decoding step to finalize the optimal architecture path (tailored for lower deployment costs and faster inference), thereby avoiding path decoding bias (particularly when the path weights are uniform). Furthermore, NAS-LoRA eliminates the re-training step, which is required to correct performance degradation caused by decoding bias in traditional NAS, further easing the training burden while preserving adaptability.

\begin{table*}[t]
  \centering
  \begin{adjustbox}{width=0.99\textwidth}
  \begin{tabular}{lccccccccccccccc}
    \toprule
    \multirow{3.5}{*}{Method} & \multirow{2}{*}{Params(M)/} &\multicolumn{6}{c}{Medical Images} & \multicolumn{4}{c}{Natural Images} & \multicolumn{2}{c}{Agriculture} & \multicolumn{2}{c}{Remote Sensing} \\
    \cmidrule(lr){3-8} \cmidrule(lr){9-12} \cmidrule(lr){13-14} \cmidrule{15-16} 
    &   & \multicolumn{2}{c}{Kvasir} & \multicolumn{2}{c}{CVC-ClinicDB} & \multicolumn{2}{c}{ISIC 2017} & \multicolumn{3}{c}{CAMO} & SBU & \multicolumn{2}{c}{Leaf}  & \multicolumn{2}{c}{Road} \\
    &  \multirow{-2}{*}{Ratio(\%)} &  $S_{\alpha}\uparrow$ & $E_{\phi}\uparrow$ & $S_{\alpha}\uparrow$ & $E_{\phi}\uparrow$ & Jac$\uparrow$ & Dice$\uparrow$ & $S_{\alpha}\uparrow$ & $E_{\phi}\uparrow$ & $F_{\beta}^{w}\uparrow$ & BER$\downarrow$ & IoU$\uparrow$ & Dice$\uparrow$ & IoU$\uparrow$ & Dice$\uparrow$ \\
    \midrule
    \multicolumn{3}{l}{\textbf{\textit{Non-PEFT methods.}}}\\
    Domain Specific & */100\% & 90.9 & 94.4 & 92.6* & 95.5* & 80.1* & 87.5* & 80.8 & 85.8 & 73.1 & 3.56 & 62.3 & 74.1 & 59.1 & 73.0 \\
    SAM (scratch) & 641.09/100\% & 78.5 & 82.4 & 85.9 & 91.6 & 73.8 & 82.5 & 61.9 & 67.0 & 40.5 & 5.53 & 52.1 & 65.5 & 55.6 & 71.1 \\
    \midrule
    \multicolumn{3}{l}{\textbf{\textit{PEFT methods.}}}\\
    Decoder-only & 3.51/0.55\% & 86.5 & 89.5 & 85.5 & 89.9 & 69.7 & 79.5 & 78.5 & 83.1 & 69.8 & 14.58 & 50.8 & 63.8 & 48.6 & 65.1 \\
    SAM-Adapter & 3.98/0.62\% & 89.6 & 92.5 & 89.6 & 92.4 & 76.1 & 84.6 & 85.6 & 89.6 & 79.8 & 3.14 & 71.4 & 82.1 & 60.6 & 75.2 \\
    Mona & 17.08/2.66\% & 91.7 & 94.4 & 91.2 & 93.9 & 77.1 & 85.3 & 87.6 & 91.6 & 82.4 & 2.72 & 74.0 & 84.1 & 61.9 & 76.1\\
    BitFit & 3.96/0.62\% & 90.8 & 93.8 & 89.0 & 91.6 & 76.4 & 84.7 & 86.8 & 90.7 & 81.5 & 3.16 & 71.4 & 81.7 & 60.6 & 75.2 \\
    VPT & 4.00/0.62\% & 91.5 & 94.3 & 91.0 & 93.7 & 76.9 & 85.1 & 87.4 & 91.4 & 82.1 & 2.70 & 73.6 & 83.8 & 60.2 & 74.9 \\
    LoRA & 4.00/0.62\% & 91.2 & 93.8 & 90.7 & 92.5 & 76.6 & 84.9 & 88.0 & 91.9 & 82.8 & 2.74 & 73.7 & 83.6 & 62.2 & 76.5 \\
    Conv-LoRA & 4.02/0.63\% & 92.0 & 94.7 & 91.3 & 94.0 & 77.6 & 85.7 & \underline{88.3} & \underline{92.4} & \underline{84.0} & 2.54 & 74.5 & 84.3 & 62.6 & 76.8 \\
    DoRA & 4.04/0.63\% & 91.8 & 94.6 & 91.4 & 94.0 & 77.2 & 85.3 & 87.3 & 91.0 & 82.2 & 3.05 & 73.8 & 83.8 & 61.3 & 75.7  \\
    VeRA & 0.32/0.05\% & 90.6 & 93.4 & 90.2 & 92.9 & 76.5 & 84.8 & 87.1 & 89.9 & 82.0 & 3.08 & 72.9 & 82.7 & 62.0 & 76.3 \\
    \midrule
    \rowcolor[HTML]{E2EFDA} 
    NAS-LoRA & 4.016/0.63\% & \underline{92.3} & \underline{94.8} & \underline{91.7} & \underline{94.3} & \textbf{78.5} &  \textbf{86.4} & 88.2 & 92.2 &  83.7 & \textbf{2.42}* & \underline{75.1} & \underline{84.8} & \underline{62.8} & \underline{77.0} \\
    \rowcolor[HTML]{E2EFDA}
    NAS-PC-LoRA & 4.009/0.62\% &  \textbf{92.4}* & \textbf{95.0}* & \textbf{91.9} & \textbf{94.5} & \underline{78.2} & \underline{86.2} & \textbf{88.4}* & \textbf{92.5}* & \textbf{84.2}* & \underline{2.52} &  \textbf{75.4}* & \textbf{84.9}* & \textbf{63.3}* & \textbf{77.4}* \\
    \bottomrule
  \end{tabular}
  \end{adjustbox}
  \vspace{-2mm}
    \caption{Results of binary-class semantic segmentation. ``Domain-Specific'' refers to methods specifically designed for the target tasks: PraNet~\cite{fan2020pranet} for Kvasir and CVC-ClinicDB, Transfuse~\cite{zhang2021transfuse} for ISIC 2017, SINet-v2~\cite{fan2021concealed} for CAMO, FDRNet~\cite{zhu2021mitigating} for SBU, Deeplabv3~\cite{chen2017deeplab} for Leaf, and LinkNet34MTL~\cite{batra2019improved} for Road. ``Params (M) / Ratio (\%)'' represents the number of trainable parameters and their proportion relative to total parameters. The best and second-best results among PEFT methods are highlighted in bold and underlined, respectively, while ``*'' indicates the best performance among both Non-PEFT and PEFT methods. Compared to existing PEFT approaches, NAS-LoRA/NAS-PC-LoRA achieves superior performance with minimal parameter overhead.}  \label{tab:compare_1}
\end{table*}

\section{Experiment}
\subsection{Experimental Setup}
\noindent \textbf{Implementation Details.}~
We apply NAS-LoRA to the $Q$, $K$ and $V$ matrices in self-attention layers of the image encoder, following the same approach as LoRA. Experiments are conducted on an Nvidia V100 GPU with a batch size of 4. For weight parameters $w$, we use the Adam optimizer with an initial learning rate of $1\times 10^{-4}$ and a weight decay of $1\times 10^{-4}$. For architecture parameters $\alpha$, the Adam optimizer is used with an initial learning rate of $1\times 10^{-3}$ and a weight decay of $1\times 10^{-3}$. The initial values of $\alpha$ before softmax are sampled from a standard Gaussian scaled by 0.001. We set the optimization steps to $T=40$ and $T_{B}=10$. Random horizontal flipping is employed as a data argumentation technique during training. The weight for the segmentation loss, $\lambda_{seg}$, is set to $1$, while the weight for the classification loss, $\lambda_{cls}$, is set to $2$. The default rank of NAS-LoRA is $3$, and the selected feature channel ratio is set to $2/3$. To ensure result stability, all experiments are conducted three times, with mean values reported.

\noindent \textbf{Baselines.}~
We compare NAS-LoRA with following methods: 1) Fine-tuning SAM's mask decoder only. 2) SAM-Adapter~\cite{chen2023sam}, which incorporates domain-specific information into SAM using adapters. 3) BitFit~\cite{zaken2021bitfit}, which fine-tunes only the bias terms of the model. 4) VPT~\cite{jia2022visual}, which inserts learnable tokens into the hidden states of each transformer layer. 5) LoRA~\cite{hu2022lora}. 6) Conv-LoRA~\cite{zhong2024convolution}, which introduces the convolution managed by Mixture-of-Expert (MoE) into the LoRA. 7) DoRA~\cite{liu2024dora}, which decomposes the pre-trained weights into the magnitude and direction for fine-tuning. 8) VeRA~\cite{kopiczko2023vera}, which shares a single pair of low-rank matrices across all layers and learns small scaling vectors instead of LoRA. 9) Mona~\cite{yin20255}, which introduces multiple vision-friendly filters into the adapter. The experimental results of these baselines are either cited from the original papers or re-implemented.


\noindent \textbf{Datasets.}~
Our experiments encompass semantic segmentation datasets from multiple domains~\cite{zhong2024convolution}, including natural images, medical images, agriculture, and remote sensing. In the medical domain, we study polyp segmentation using CVC-ClinicDB~\cite{bernal2015wm} and Kvasir datasets~\cite{jha2020kvasir}, and skin lesion segmentation with ISIC 2017~\cite{codella2018skin}. For natural images, we investigate camouflaged object segmentation on COD10K~\cite{fan2020camouflaged}, CAMO~\cite{le2019anabranch}, and CHAMELEON~\cite{skurowski2018animal}, and shadow detection on SBU~\cite{vicente2016large}. In agriculture, we evaluate on the Leaf Disease Segmentation dataset~\cite{rath2023leaf}, and for remote sensing, we use the Massachusetts Roads dataset~\cite{mnih2013machine}. Additionally, we explore multi-class segmentation on transparent object datasets, Trans10K-v1 (3 classes)~\cite{xie2020segmenting} and Trans10K-v2 (12 classes)~\cite{xie2021segmenting}. Further dataset details are provided in Appx.~\ref{sec:app_dataset}.

\begin{table*}[ht]
  \centering
    \begin{adjustbox}{width=0.9\textwidth}
  \begin{tabular}{lccccccccc}
    \toprule
    \multirow{2}{*}{Method} & \multirow{2}{*}{Params(M)/Ratio(\%)}  & \multicolumn{4}{c}{Easy} & \multicolumn{4}{c}{Hard} \\
    &  & Acc$\uparrow$ & mIoU$\uparrow$ & MAE$\downarrow$ & MBER$\uparrow$ & Acc$\uparrow$ & mIoU$\uparrow$ & MAE$\downarrow$ & MBER$\downarrow$ \\
    \midrule
    \multicolumn{2}{l}{\textbf{\textit{Non-PEFT methods.}}}\\
    Translab~\cite{xie2020segmenting} & 42.19/100\% & 95.77 & 92.23 & 0.036 & 3.12 & 83.04 & 72.10 & 0.166 & 13.30 \\
    \midrule
    \multicolumn{2}{l}{\textbf{\textit{PEFT methods.}}}\\
    Decoder-only & 3.51/0.55\% & 94.68 & 88.54 & 0.050 & 4.24 & 83.53 & 68.30 & 0.186 & 14.37 \\
    VPT & 4.00/0.62\% & 98.31 & 95.73 & 0.017 & 1.52 & 90.42 & 83.38 & 0.083 & 7.21 \\
    LoRA & 4.00/0.62\% & 98.44 & 96.26 & 0.016 & 1.35 & 91.94 & 83.95 & 0.083 & 6.35 \\
    Conv-LoRA & 4.02/0.63\% & 98.63 & 96.45 & 0.015 & 1.27 & 93.05 & 84.37 & 0.075 & 6.25 \\
    DoRA & 4.04/0.63\% & 98.52 & 96.31 & 0.016 & 1.32 & 92.21 & 84.13 & 0.078 & 6.32 \\
    VeRA & 0.32/0.05\% & 98.14 & 95.67 & 0.019 & 1.58 & 90.23 & 83.12 & 0.084 & 7.46 \\
    \midrule
    \rowcolor[HTML]{E2EFDA}
    NAS-LoRA & 4.016/0.63\% & \underline{98.74} & \underline{96.56} & \underline{0.014} & \underline{1.22} & \underline{93.82} & \underline{84.64} & \underline{0.072} & \underline{6.19} \\
    \rowcolor[HTML]{E2EFDA}
    NAS-PC-LoRA & 4.009/0.62\% & \textbf{98.78} & \textbf{96.60} & \textbf{0.013} & \textbf{1.19} & \textbf{94.00} & \textbf{84.78} & \textbf{0.070} & \textbf{6.16} \\
    \bottomrule
  \end{tabular}
  \end{adjustbox}
  \vspace{-2mm}
  \caption{Results of multi-class semantic segmentation on the Trans10K-v1 dataset (three-class segmentation). The best and second-best results are highlighted in bold and underlined, respectively. Compared to existing non-PEFT and PEFT methods, NAS-LoRA/NAS-PC-LoRA achieves superior segmentation performance with minimal parameter overhead.}
  \label{tab:compare_2}
\end{table*}

\begin{table*}[ht]
  \centering
  \begin{adjustbox}{width=0.99\textwidth}
  \begin{tabular}{lccccccccccccccc}
    \toprule
    \multirow{2}{*}{Method} & {Params(M)/}   & \multicolumn{12}{c}{Category IoU$\uparrow$} & \multirow{2}{*}{Acc$\uparrow$} & \multirow{2}{*}{mIoU$\uparrow$} \\ \cmidrule{3-14}
    & Ratio(\%) &  bg & shelf & jar & freezer & window & door & eyeglass & cup & wall & bowl & bottle & box &  &  \\
    \midrule
    \multicolumn{3}{l}{\textbf{\textit{Non-PEFT methods.}}}\\
    Translab~\cite{xie2021segmenting} & 42.19/100\%  & 93.90 & 54.36* & 64.48 & 65.14* & 54.58 & 57.72 & 79.85 &  81.61 &  72.82 &  69.63 &  77.50 & 56.43 & 92.67 & 69.00 \\
    Trans2Seg~\cite{xie2021segmenting} & 56.20/100\% & 95.35 & 53.43 & 67.82* & 64.20 & 59.64* & 60.56 & 88.52* &  86.67* &  75.99 &  73.98* & 82.43* & 57.17* & 94.14 & 72.15* \\
    \midrule
    \multicolumn{3}{l}{\textbf{\textit{PEFT methods.}}}\\
    Decoder-only & 3.51/0.55\%  & 93.66 & 32.75 & 39.96 & 35.87 & 50.70 & 45.89 & 57.38 & 73.16 & 69.36 & 54.23 & 56.58 & 33.77 & 90.66 & 49.97\\
    VPT & 4.00/0.62\% & 97.41 & 29.76 & 52.82 & 62.09 & \underline{55.54} & 63.61 & 81.12 & 83.40 & 79.61 & 65.29 & 72.92 & 44.77 & 94.42 & 62.81 \\
    LoRA & 4.00/0.62\% & 97.50 & 42.17 & 57.82 & 64.35 & 53.44 & 64.08 & 87.28 & 85.28 & 80.43 & 63.67 & 77.97 & 49.56 & 94.80 & 66.01 \\
    Conv-LoRA & 4.02/0.63\% & 97.66 & \underline{50.51} & 58.44 & 51.70 & \textbf{55.69} & 65.22 & 85.23 & 84.84 & 80.97 & \textbf{72.84} & \textbf{79.83} & \underline{52.73} & 95.07 & 67.09 \\
    DoRA & 4.04/0.63\% & 97.58 & 45.29 & 58.12 & 60.45 & 54.23 & 64.54 & 86.12 & 85.12 & 80.23 & 64.67 & \underline{78.97} & 51.56 & 94.97 & 66.67 \\
    VeRA & 0.32/0.05\% & 97.33 & 30.12 & 53.21 & 61.08 & 54.53 & 63.47 & 82.36 & 83.95 & 80.10 & 64.43 & 75.84 & 46.62 & 94.63 & 64.46 \\
    \midrule
    \rowcolor[HTML]{E2EFDA}
    NAS-LoRA & 4.016/0.63\% & \underline{97.78} & {49.90} & \underline{59.24} & \textbf{65.06} & {54.83} & \underline{65.77} & \underline{87.86} & \underline{85.71} & \underline{81.04} & {68.41} &  {78.59} & \textbf{52.82} & \underline{95.28} & \underline{67.86} \\
    \rowcolor[HTML]{E2EFDA}
    NAS-PC-LoRA & 4.009/0.62\% & \textbf{97.80}* & \textbf{50.71} & \textbf{59.42} & \underline{64.88} & {55.01} & \textbf{65.98}* & \textbf{88.23} & \textbf{85.91} & \textbf{81.35}* & \underline{69.45} & {78.78} & {52.08} & \textbf{95.31}* & \textbf{67.94} \\
    \bottomrule
  \end{tabular}
  \end{adjustbox}
  \vspace{-2mm}
  \caption{Results of multi-class semantic segmentation on the Trans10K-v2 dataset (twelve-class segmentation). ``*'' denotes the best overall results across both non-PEFT and PEFT methods. Compared to existing approaches, NAS-LoRA/NAS-PC-LoRA achieves superior segmentation performance with minimal parameter overhead.}
  \label{tab:compare_3}
\end{table*}

\begin{figure}[ht]
  \centering
  \includegraphics[width=0.99\linewidth]{Figures/visulization.pdf}
  \vspace{-3mm}
  \caption{Visual comparisons on sample images from the ISIC 2017 ($1^{st}$ line), Leaf ($2^{nd}$ line), CAMO ($3^{rd}$ line), Road ($4^{th}$ line), and Transparent Object ($5^{th}$ line) datasets.}
  \label{fig:visualization}
\end{figure}

\vspace{-2mm}
\subsection{Main Results}

\noindent \textbf{Binary-Class Semantic Segmentation.}
Table~\ref{tab:compare_1} presents the results of binary-class semantic segmentation across each datasets. NAS-LoRA and NAS-PC-LoRA outperform the baselines on most datasets, highlighting the effectiveness of optimizing priors with NAS techniques and the importance of inductive bias in visual tasks.
In terms of parameter efficiency, our methods achieve superior performance with minimal parameter overhead. While NAS-LoRA falls slightly behind domain-specific methods on certain tasks, its strong generalization ability allows it to be easily adapted to various downstream tasks without requiring domain-specific information.
Additionally, NAS-PC-LoRA consistently outperforms NAS-LoRA on most datasets, indicating the effectiveness of the partial connection mechanism in the PEFT setting. A possible explanation is that partial channel sampling reduces bias in operation selection and regularizes the preference for weight-free operations~\cite{xu2019pc}, which improves the stability and performance of NAS.

\noindent \textbf{Multi-Class Semantic Segmentation.}
Table~\ref{tab:compare_2} and Table~\ref{tab:compare_3} present the results of multi-class semantic segmentation on Trans10K-v1 and Trans10K-v2. For Trans10K-v1, our proposed method outperforms existing non-PEFT and PEFT approaches across both Easy and Hard attributes. 
On Trans10K-v2, NAS-LoRA/NAS-PC-LoRA achieve the highest category IoU in nine categories among PEFT methods and attain the highest overall accuracy compared to all other methods.
Notably, existing PEFT approaches underperform compared to non-PEFT methods in mIoU, as SAM’s image encoder struggles to extract high-level semantic information crucial for classification tasks~\cite{zhong2024convolution}. Our method further bridges this gap by dynamically adjust adaptation configurations, demonstrating its ability to better retain and leverage high-level semantic information.

\begin{table}[t]
\small
\centering
\begin{adjustbox}{width=0.99\linewidth}
\begin{tabular}{l|ccc|ccc}
\toprule
\textbf{Model} & \multicolumn{3}{c|}{\textbf{Attention Distance$\downarrow$}} & \multicolumn{3}{c}{\textbf{Log Amplitude$\uparrow$}} \\
 & Layer 10 & Layer 15 & Layer 20 & $f=0.4\pi$ & $f=0.6\pi$ & $f=0.8\pi$ \\
\midrule
Randomly   & 114 & 117 & 115 & -5.0 & -5.2 & -5.4 \\
SAM ViT    & 76  & 105 & 93  & -4.2 & -4.4 & -4.5 \\
NAS-LoRA   & 65  & 98  & 82  & -4.0 & -4.3 & -4.4 \\
\bottomrule
\end{tabular}
\end{adjustbox}
\vspace{-2mm}
\caption{Evaluation of inductive biases within features.}\label{tab:inductive_bias}
\end{table}

\noindent \textbf{Visualization Analysis.}~
Fig.~\ref{fig:visualization} presents comparisons of various representative methods across the ISIC 2017, Leaf, CAMO, Road, and Transparent Object datasets. The results demonstrate that NAS-LoRA consistently produces more precise segmentation results.

\begin{figure}[t]
  \centering
  \includegraphics[width=0.99\linewidth]{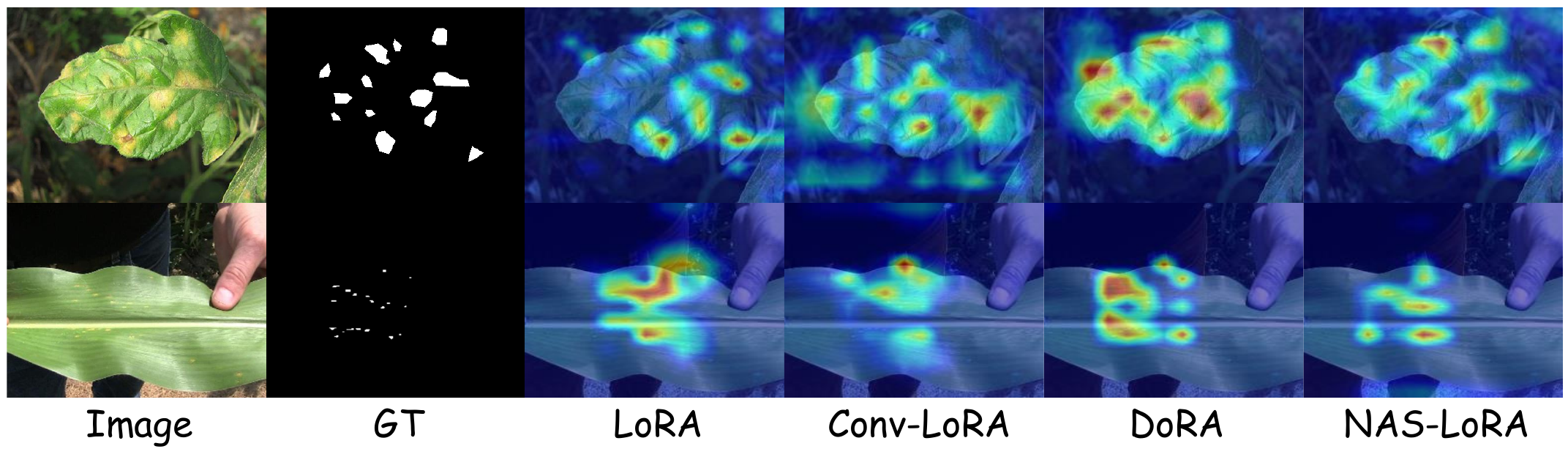}
  \vspace{-3mm}
  \caption{Heatmap visualization by Grad-CAM~\cite{selvaraju2017grad} on Leaf. Compared to previous methods, applying NAS-LoRA could capture more fine-grained details.}
  \label{fig:visualization_feature}
\end{figure}

\noindent\textbf{Enhanced Inductive Bias.}~
We use mean attention distance~\cite{raghu2021vision} and relative log amplitude~\cite{park2022vision} to evaluate the inductive biases of the model on 100 randomly sampled segmentation images. As shown in the Table~\ref{tab:inductive_bias}, from randomly initialized ViT to SAM ViT and ultimately to NAS-LoRA, the mean attention distance decreases, while the high-frequency signals in the features increase, illustrating a focus on local information and the enhancement of inductive bias by NAS-LoRA. Fig.~\ref{fig:visualization_feature} visualizes heatmaps from SAM' image encoder by Grad-CAM~\cite{selvaraju2017grad}, where NAS-LoRA could provide more accurate information, which is beneficial to later mask prediction, showing the enhanced inductive bias.

\begin{table}[ht] \small
  \centering
    \begin{adjustbox}{width=0.99\linewidth}
  \begin{tabular}{lccc}
    \toprule
    \multirow{2}{*}{Method} & \multicolumn{2}{c}{Training} & Inference \\
    & iter/s$\uparrow$ & min/epoch$\downarrow$ & iter/s$\uparrow$ \\
    \midrule
    BitFit & 1.82 & 18.3 & 4.62 \\
    Adapter & 1.73 & 19.3 & 4.38 \\
    VPT & 1.41 & 23.6 & 4.39 \\
    LoRA & 1.64 & 20.3 & 4.62 \\
    Conv-LoRA & 1.22 & 27.3 & 3.64 \\
    DoRA & 1.61 & 20.7 & 4.62 \\
    VeRA &  13.14 & 2.54 & 4.62 \\
    Ours (update $w$) & 1.57 & 21.1 & \multirow{2}{*}{4.62}\\
    Ours (update $w\&\alpha$) & - & 25.2 &  \\
    \bottomrule
  \end{tabular}
  \end{adjustbox}
  \vspace{-2mm}
    \caption{Computational cost comparison of NAS-PC-LoRA and other PEFT methods on the ISIC 2017 dataset.}
  \label{tab:cost}
\end{table}
\vspace{-5mm}
\begin{table}[ht]
  \centering
  \begin{adjustbox}{width=\linewidth}
  \begin{tabular}{lcccc}
    \toprule
    \multirow{2}{*}{Method} & \multicolumn{4}{c}{ISIC 2017} \\
    & Jac$\uparrow$ & T-Jac$\uparrow$ & Dice$\uparrow$ & Acc$\uparrow$ \\
    \midrule
    NAS-PC-LoRA & 78.2 & 71.2 & 86.2 & 94.6\\
    w/o convolution & 76.9$_{\textcolor{red}{1.3\downarrow}}$ & 68.6$_{\textcolor{red}{2.6\downarrow}}$ & 85.0$_{\textcolor{red}{1.2\downarrow}}$ & 93.3$_{\textcolor{red}{1.3\downarrow}}$ \\
    w/o pooling & 77.9$_{\textcolor{red}{0.3\downarrow}}$ & 70.5$_{\textcolor{red}{0.7\downarrow}}$ & 85.9$_{\textcolor{red}{0.3\downarrow}}$ & 94.2$_{\textcolor{red}{0.4\downarrow}}$ \\
    w/o skip/no connection & 77.5$_{\textcolor{red}{0.7\downarrow}}$ & 69.7$_{\textcolor{red}{1.5\downarrow}}$ & 85.6$_{\textcolor{red}{0.6\downarrow}}$ & 94.0$_{\textcolor{red}{0.6\downarrow}}$ \\
    \bottomrule
  \end{tabular}
\end{adjustbox}
\vspace{-2mm}
  \caption{Effects of candidate operations in the NAS cell.}
  \label{tab:ablation_candidate}
\end{table}
\vspace{-2mm}
\noindent \textbf{Computational Cost.}~
In Table~\ref{tab:cost}, we compare the computational cost of NAS-PC-LoRA with other PEFT methods in terms of training and inference. The evaluation is conducted on ISIC 2017 using an Nvidia V100 GPU. While NAS-PC-LoRA has a higher training cost than the original LoRA due to its stage-wise optimization strategy, it introduces no additional inference overhead and delivers robust performance gains across various tasks, as shown in Tables~\ref{tab:compare_1},~\ref{tab:compare_2}, and~\ref{tab:compare_3}.

\vspace{-2mm}
\subsection{Ablation Study}
\noindent \textbf{Effect of Operation.}~
Table~\ref{tab:ablation_candidate} presents the effects of different candidate operations in the NAS cell. Convolution contributes the most to performance, followed by skip/no connection, while pooling has a relatively smaller impact.

\begin{table}[ht]
  \centering
    \begin{adjustbox}{width=0.8\linewidth}
  \begin{tabular}{lccc}
    \toprule
    \multirow{2}{*}{LoRA} & \multirow{2}{*}{Params(M)/Ratio(\%)} & \multicolumn{2}{c}{Trans10K-v2} \\
    &  & Acc$\uparrow$ & mIoU$\uparrow$   \\
    \midrule
    $r=3$ & 4.009/0.62\% & 95.31 & 67.94 \\
    $r=6$ & 4.50/0.70\% & 95.72 & 68.36  \\
    $r=12$ & 5.49/0.86\% & \textbf{95.77} & 68.48 \\
    $r=24$ & 7.46/1.16\% & 95.76 & \textbf{68.65} \\
    \bottomrule
  \end{tabular}
  \end{adjustbox}
  \vspace{-2mm}
    \caption{Effects of the rank $r$ of NAS-LoRA.}
  \label{tab:ablation_rank}
\end{table}
\vspace{-2mm}
\noindent \textbf{Rank of LoRA.}~
We further investigate the effects of the rank $r$ of NAS-LoRA in Table~\ref{tab:ablation_rank}. The performance increases with the increase of the rank $r$, while the high rank also causing the parameter overhead and disobeying the principle of parameter efficiency of PEFT.

\begin{table}[ht]
  \centering
    \begin{adjustbox}{width=0.99\linewidth}
  \begin{tabular}{lcccc}
    \toprule
    \multirow{2}{*}{LoRA Blocks} & {Params(M)/} & \multicolumn{3}{c}{Leaf} \\
    & Ratio(\%) & IoU$\uparrow$ & Dice$\uparrow$ & Acc$\uparrow$ \\
    \midrule
    Layers \{1,2,3,\dots,16\} & 2.004/0.31\% & 73.0 & 82.9 & 95.5\\
    Layers \{16,17,18,\dots,32\} & 2.004/0.31\% & 73.6 & 83.4 & 95.7 \\
    Layers \{1,2,3,\dots,32\} & 4.009/0.62\% & \textbf{75.4} & \textbf{84.9} & \textbf{96.6} \\  
    \bottomrule
  \end{tabular}
  \end{adjustbox}
    \vspace{-2mm}
    \caption{Ablation study on the place of NAS blocks in ViTs.}
  \label{tab:ablation_layers}
\end{table}
\vspace{-2mm}
\noindent \textbf{Effect of LoRA Place.}~
The effect of the NAS-LoRA place in ViT layers is shown in the Table~\ref{tab:ablation_layers}. While the performance is optimal when NAS blocks are inserted into all layers, we find that inserting them into the last half of the layers performs better than inserting them into the first half.

\begin{table}[ht]
  \centering
  \begin{adjustbox}{width=0.99\linewidth}
  \begin{tabular}{lcccc}
    \toprule
    \multirow{2}{*}{Architecture Epoch $T_{B}$} & \multicolumn{4}{c}{ISIC 2017} \\
    & Jac$\uparrow$ & T-Jac$\uparrow$ &  Dice$\uparrow$ & Acc$\uparrow$ \\
    \midrule
    0 (update $w\&\alpha$ simultaneously) & 77.5 & 70.1 & 85.3 & 93.9 \\
    0 (update $w\&\alpha$ stage-wisely) & 77.9 & 70.6 & 85.8 & 94.4 \\
    \rowcolor[HTML]{E2EFDA}
    10 (ours) & \textbf{78.2} & \textbf{71.2} & \textbf{86.2} & \textbf{94.6} \\
    20  & 78.0 & 70.9 & 85.9 & 94.4 \\
    \bottomrule
  \end{tabular}
  \end{adjustbox}
    \vspace{-1.7mm}
    \caption{Effects of the stage-wise optimization strategy.}
  \label{tab:ablation_balance}
\end{table}
\vspace{-2mm}
\noindent \textbf{Effect of Stage-wise Optimization.}~
We assess the effect of the stage-wise optimization in Table~\ref{tab:ablation_balance}. When architecture parameters are optimized together with the weights, performance degrades significantly. 
In contrast, stage-wise optimization—where architecture parameters are updated after weights have partially converged—yields better results. Although early and delayed architecture optimization affect performance, the impact remains within an acceptable range.

\vspace{-2mm}
\section{Conclusion}
In this paper, we propose NAS-LoRA, a PEFT-based segmentation approach that integrates NAS to optimize local priors. Through an end-to-end stage-wise optimization strategy, NAS-LoRA effectively adapts to downstream tasks while maintaining minimal training parameter overhead and incurring no additional inference cost. Experiments across various segmentation tasks validate its effectiveness.

\bibliography{aaai2026}

\newcommand{\answerYes}[1][]{\textcolor{blue}{[Yes] #1}}
\newcommand{\answerNo}[1][]{\textcolor{orange}{[No] #1}}
\newcommand{\answerNA}[1][]{\textcolor{gray}{[N/A] #1}}

\clearpage
\newpage
\renewcommand{\thesection}{\Alph{section}}
\renewcommand\thefigure{\Alph{section}\arabic{figure}} 
\renewcommand\thetable{\Alph{section}\arabic{table}}  
\setcounter{section}{0}
\setcounter{figure}{0} 
\setcounter{table}{0} 

\appendix

In the supplementary material, we provide additional details and experimental results to enhance the understanding and insights into our proposed NAS-LoRA. This supplementary material is organized as:
\begin{itemize}[leftmargin=*]

\item In Sec.~\ref{sec:app_loss}, we provide more details about the loss functions in the training.

\item In Sec.~\ref{sec:app_dataset}, we provide more details about the employed datasets.

\item In Sec.~\ref{sec:app_train}, we provide more details about the training behavior, including the training convergence analysis and the adaptability of NAS-LoRA.

\item In Sec.~\ref{sec:app_ablation}, we provide more ablation studies, including the effect of the partial connection in the NAS-PC-LoRA. 

\end{itemize}

\section{Loss Functions}\label{sec:app_loss}

Our final loss function combines the segmentation loss $\mathcal{L}_{\text{seg}}$ and classification loss $\mathcal{L}_{\text{cls}}$ as follows: 
\begin{equation} 
\mathcal{L} = \lambda_{\text{seg}} \mathcal{L}_{\text{seg}} + \lambda_{\text{cls}} \mathcal{L}_{\text{cls}}, 
\end{equation} 
where $\lambda_{\text{seg}}$ and $\lambda_{\text{cls}}$ are the weighting coefficients that balance the two objectives.
The segmentation loss $\mathcal{L}_{\text{seg}}$~\cite{ma2024segment} includes a binary cross-entropy (BCE) loss $\mathcal{L}_{\text{BCE}}$ and a Dice loss $\mathcal{L}_{\text{Dice}}$, defined respectively as:
\vspace{-0.8mm}
\begin{equation} 
\mathcal{L}_{\text{BCE}} = -\frac{1}{N} \sum_{i=1}^{N} \left[y_i \log(\hat{y}_i) + (1 - y_i) \log(1 - \hat{y}_i)\right], 
\end{equation} 
\begin{equation} 
\mathcal{L}_{\text{Dice}} = 1 - \frac{2 \sum_{i=1}^{N} y_i \hat{y}_i}{\sum_{i=1}^{N} y_i^2 + \sum_{i=1}^{N} \hat{y}_i^2}. 
\end{equation}
Here, $N$ is the total number of pixels, $y_i$ is the ground-truth label, and $\hat{y}_i$ is the predicted segmentation output.

The classification loss $\mathcal{L}_{\text{cls}}$ is defined as the cross-entropy over all $K+1$ classes: 
$
\mathcal{L}_{\text{cls}} = -\frac{1}{N} \sum_{i=1}^{N} \sum_{k=1}^{K+1} y_i^k \log(\hat{y}_i^k), 
$
where $y_i^k$ and $\hat{y}_i^k$ denote the ground-truth and predicted probability for class $k$ at pixel $i$, respectively.

\section{Dataset Details}\label{sec:app_dataset}

\noindent\textbf{Medical Images.}~
The polyp segmentation task involves segmenting abnormal growths in gastrointestinal endoscopy images, which is challenging due to complex backgrounds and the diverse sizes and shapes of polyps. We use two polyp segmentation datasets: CVC-ClinicDB~\cite{bernal2015wm} (612 images) and Kvasir~\cite{jha2020kvasir} (1000 images). Following~\cite{fan2020pranet,zhong2024convolution}, we split each dataset into 90\% for training and 10\% for testing, with 20\% of the training set used for validation. Additionally, we evaluate our method on the skin lesion segmentation task, which involves segmenting lesions from dermoscopic images. For this, we use the ISIC 2017 dataset~\cite{codella2018skin}, comprising 2000 training images, 150 validation images, and 600 testing images.

\noindent\textbf{Natural Images.}~
Camouflaged object segmentation aims to segment objects concealed within complex or visually cluttered backgrounds. We use three related datasets: COD10K~\cite{fan2020camouflaged}, CAMO~\cite{le2019anabranch}, and CHAMELEON~\cite{skurowski2018animal}. COD10K consists of 3,040 training and 2,026 testing images, CAMO includes 1,000 training and 250 testing images, while CHAMELEON provides 76 images for testing. The model is trained on the training sets of COD10K and CAMO and evaluated on all three datasets, with CAMO highlighted as the most challenging.
Additionally, we conduct experiments on the Shadow Detection dataset, SBU~\cite{vicente2016large}, which focuses on identifying shadow regions within a scene. SBU contains 4,085 training and 638 testing images. For both tasks, 10\% of the training set is randomly selected as the validation set.

\noindent\textbf{Agriculture.}~
Leaf Disease Segmentation dataset~\cite{rath2023leaf} is used to identify the infected regions on the leaves. The dataset includes 498 images for training and 90 images for testing.

\noindent\textbf{Remote Sensing.}~
Road segmentation aims to identify road or street regions in images. We use the Massachusetts Roads Dataset~\cite{mnih2013machine}, which includes 1,107 training images, 13 validation images, and 48 testing images.

\noindent\textbf{Multi-class Segmentation.}~
For a more comprehensive evaluation, we also explore multi-class segmentation using the multi-class transparent object segmentation datasets Trans10K-v1~\cite{xie2020segmenting}, which includes three classes (background and two categories of transparent objects), and Trans10K-v2~\cite{xie2021segmenting}, which consists of 12 classes (background and 11 fine-grained categories of transparent objects). Both datasets contain 5,000 training images, 1,000 validation images, and 4,428 testing images.

\section{Training Behavior}\label{sec:app_train} 
\begin{figure}[ht]
  \centering
  \includegraphics[width=\linewidth]{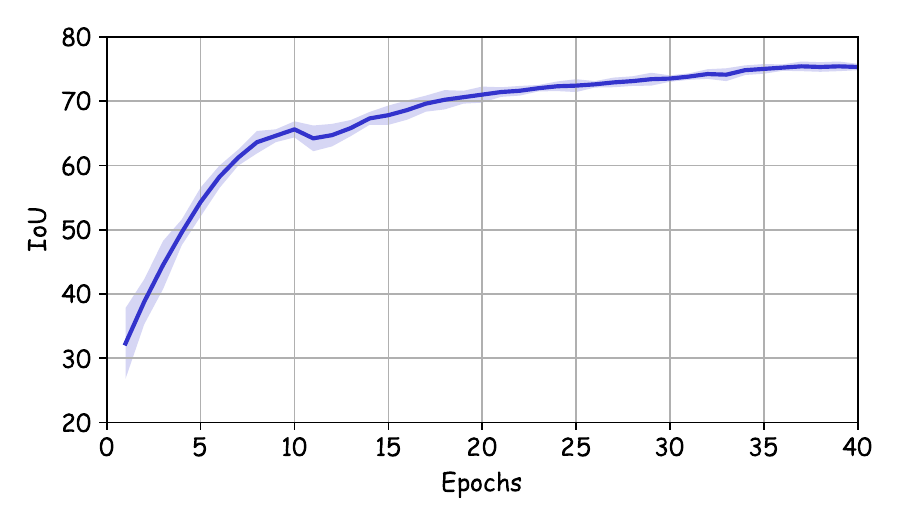}
  \caption{Convergence behavior of NAS-PC-LoRA on the Leaf dataset over three independent trials.}
  \label{fig:convergence}
\end{figure}

\noindent\textbf{Convergence Analysis.}~
A key concern when integrating NAS with PEFT is ensuring stable convergence. Fig.~\ref{fig:convergence} illustrates the convergence behavior of NAS-PC-LoRA on the Leaf dataset across three independent trials, where the dark blue line denotes the mean value, and the light blue shaded area represents the standard deviation. The curve exhibits a rapid initial increase, followed by a more gradual rise before stabilizing. Notably, a slight fluctuation occurs around the 10-$th$ epoch due to the optimization of architectural parameters, after which the trend resumes its upward trajectory. These results confirm that NAS-PC-LoRA effectively converges to a stable solution.

\begin{figure}
  \centering
  \includegraphics[width=\linewidth]{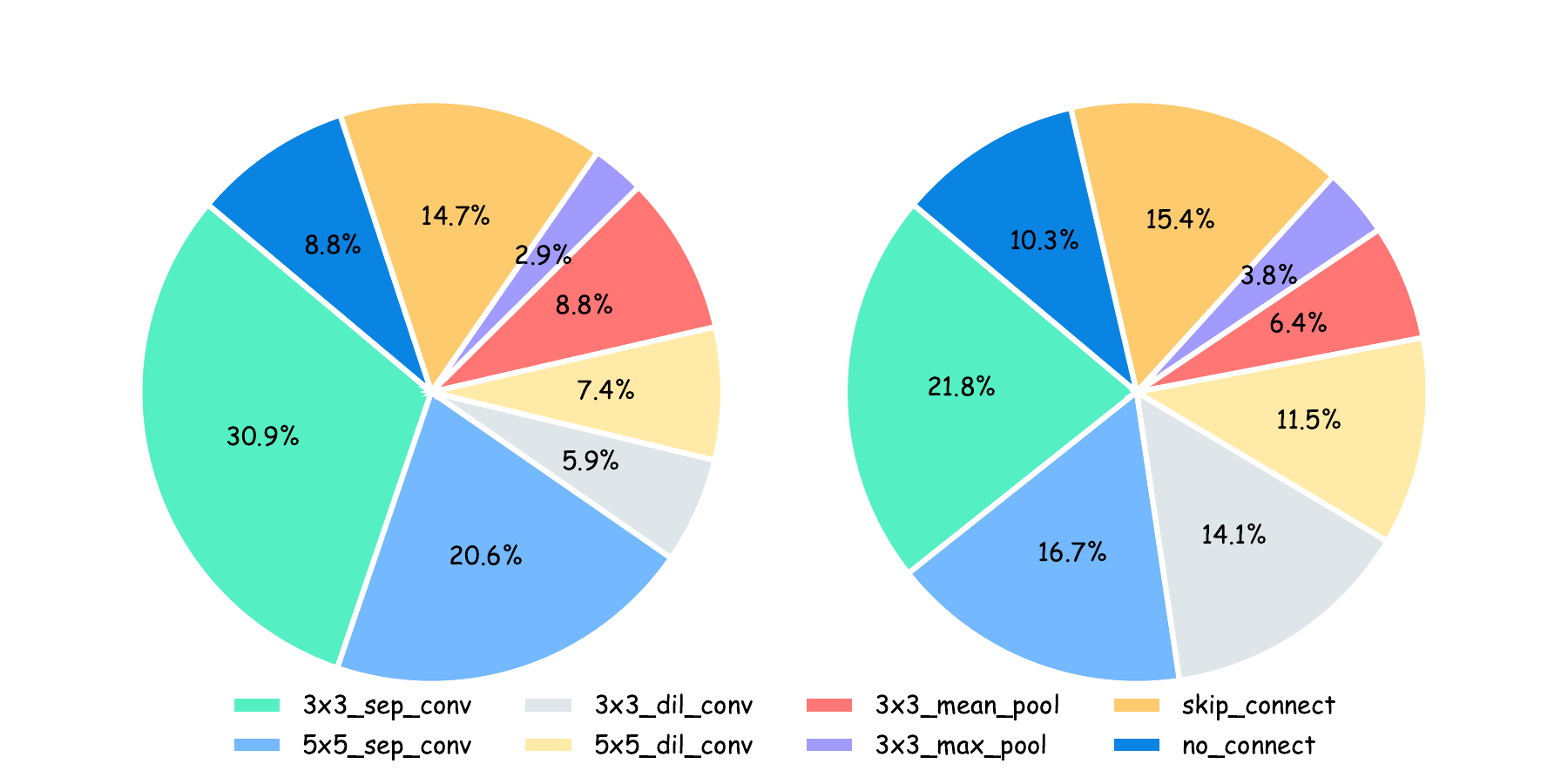}
  \caption{The final architecture of NAS-PC-LoRA on Leaf Disease Segmentation dataset (Left) and Multi-class Transparent Object Segmentation dataset (Right). }
  \label{fig:operation}
\end{figure}

\noindent\textbf{Adaptability of NAS-PC-LoRA.}~
To better understand the adaptability of NAS-PC-LoRA across different downstream tasks, we analyze the statistical distribution of operation selections on the Leaf Disease Segmentation and Multi-class Transparent Object Segmentation datasets, as shown in Fig.~\ref{fig:operation}. The proportion of each candidate operation is computed as: 
$
    P_{O_{i}\in\mathcal{O}} = \frac{\sum_{layers}\sum_{q,k,v}\alpha_{i}}{\sum layers\sum q,k,v}
$. The results reveal distinct operation preferences across the two datasets: the Leaf dataset predominantly selects separable convolution, while the Transparent Object dataset favors both dilated and separable convolutions. This diversity in operation selection highlights its ability to adapt to the unique characteristics of different tasks.

\section{More Ablation Studies}\label{sec:app_ablation}

\begin{table}[ht]
  \centering
    \begin{adjustbox}{width=0.99\linewidth}
  \begin{tabular}{lcccc}
    \toprule
    \multirow{2}{*}{Sampling Channels} & {Params(M)/} & \multicolumn{3}{c}{Leaf} \\
    & Ratio(\%) & IoU$\uparrow$ & Dice$\uparrow$ & Acc$\uparrow$ \\
    \midrule
    3/3 (NAS-LoRA) & 4.016/0.63\% & 75.1 & 84.8 & 96.4\\
    2/3 (NAS-PC-LoRA) & 4.009/0.62\% & \textbf{75.4} & \textbf{84.9} & \textbf{96.6} \\
    1/3 & 4.003/0.62\% & 74.8 & 84.6 & 96.2 \\  
    \bottomrule
  \end{tabular}
  \end{adjustbox}
    \vspace{-1.7mm}
    \caption{Effects of the partially connection in NAS-LoRA.}
  \label{tab:ablation_pc}
\end{table}

\noindent \textbf{Partial Connection.}~
A notable feature of NAS-PC-LoRA is its partial connection mechanism, which allows for the update of a subset of feature channels. In Table~\ref{tab:ablation_pc}, we show the effects of partial connection, indicating that this setting is meaningful in the PEFT scenario. The optimal sampling ratio is 2/3, as too few or too many channels may lead to performance degradation.

\end{document}